\documentclass[11pt]{article}

\usepackage[preprint]{acl}

\usepackage{times}
\usepackage{latexsym}

\usepackage[T1]{fontenc}

\usepackage[utf8]{inputenc}

\usepackage{microtype}

\usepackage{inconsolata}

\usepackage{graphicx}

\usepackage{amsmath, amssymb}
\usepackage{booktabs,multirow}
\usepackage{graphicx}
\usepackage{subcaption}
\usepackage[most]{tcolorbox}
\usepackage{tabularx}
\usepackage{ragged2e}

\newcommand{\tightlist}{%
  \setlength{\itemsep}{0pt}%
  \setlength{\parskip}{0pt}%
}

\usepackage{array}  

\newcommand{\grouphead}[1]{%
  \multicolumn{4}{c}{%
    \parbox[c]{3.0cm}{\centering #1}%
  }%
}

\newcommand{\framework}{\textsc{GradFiltering}}

\title{Uncertainty-Aware Gradient Signal-to-Noise Data Selection for \\Instruction Tuning}

\author{
 \textbf{Zhihang Yuan\textsuperscript{1}},
 \textbf{Chengyu Yue\textsuperscript{1}},
 \textbf{Long Huang\textsuperscript{1}},
 \textbf{Litu Ou\textsuperscript{2}},
 \textbf{Lei Shi\textsuperscript{1,}
 \thanks{Corresponding Author. \href{juetian.sl@alibaba-inc.com}{juetian.sl@alibaba-inc.com}}}
\\
\\
 \textsuperscript{1}Alibaba Cloud Computing
 \textsuperscript{2}The University of Edinburgh
\\
\textsuperscript{1}\{yuanzhihang.yzh,yuechengyu.ycy, baixuan.hl, juetian.sl\}@alibaba-inc.com
\\
\textsuperscript{2}litu.ou@ed.ac.uk
}

\begin{document}
\maketitle

\begin{abstract}
Instruction tuning is a standard paradigm for adapting large language models (LLMs), but modern instruction datasets are large, noisy, and redundant, making full-data fine-tuning costly and often unnecessary.
Existing data selection methods either build expensive gradient datastores or assign static scores from a weak proxy, largely ignoring evolving uncertainty,and thus missing a key source of LLM interpretability.
We propose \framework, an objective-agnostic, uncertainty-aware data selection framework that utilizes a small GPT-2 proxy with a LoRA ensemble and aggregates per-example gradients into a Gradient Signal-to-Noise Ratio (G-SNR) utility.
Our method matches or surpasses random subsets and strong baselines in most LLM-as-a-judge evaluations as well as in human assessment.
Moreover, \framework-selected subsets converge faster than competitive filters under the same compute budget, reflecting the benefit of uncertainty-aware scoring. 
\end{abstract}

\section{Introduction}
Instruction tuning has become a standard recipe for adapting large language models (LLMs) to follow human instructions~\citep{ouyang2022training}.
Modern instruction datasets can contain hundreds of thousands of examples~\citep{taori2023alpaca,xu2023wizardlm,wang2023self,xu2024survey}, making full-data fine-tuning expensive and often unnecessary: many examples are redundant or noisy, while a relatively small subset can suffice to achieve strong performance~\citep{zhou2023lima}.
This motivates the problem of \emph{data selection} for instruction tuning: can we identify a small subset of training examples that matches or even improves the performance of full-data fine-tuning, while reducing cost?

A long line of work has studied data valuation via influence functions and related gradient-based criteria.
Classical influence-function methods estimate the effect of upweighting or removing a training point on a target loss using second-order Taylor expansions around the empirical risk minimizer~\citep{koh2017understanding}, or first-order approximations such as TracIn~\citep{pruthi2020estimating}.
While theoretically grounded, these methods are computationally expensive and brittle for deep networks, and do not scale effectively to modern LLM fine-tuning pipelines.

More recent work tailors data selection to instruction tuning.
LESS~\citep{xia2024less} constructs a gradient datastore with low-rank representations to enable optimizer-aware similarity search.
Superfiltering~\citep{li2024superfiltering} adopts a weak-to-strong strategy, using a small proxy model to filter a top-$k$ subset based on Instruction-Following Difficulty (IFD) scores.
These approaches achieve strong performance, but also have limitations: LESS requires computing and storing per-example gradients of a strong model and is inherently constrained by its dependency on task-specific validation sets, while Superfiltering assigns each example a \emph{static} difficulty score from a pre-trained proxy, without modeling interactions between training samples or the evolution of uncertainty during fine-tuning.

In parallel, epistemic uncertainty has emerged as a useful lens on data quality.
Large-scale instruction datasets inevitably contain noisy, out-of-distribution, or spurious examples. Uncertainty helps detect such samples by providing a signal that is complementary to training dynamics~\citep{gal2016dropout, lakshminarayanan2017simple}.
LoRA-Ensemble~\citep{muhlematter2024lora} shows that training multiple LoRA heads on a shared backbone can recover much of the accuracy and calibration benefits of deep ensembles at a fraction of the parameters and compute cost, suggesting that lightweight LoRA ensembles can serve as practical uncertainty probes.
However, these ideas have not yet been leveraged for uncertainty-aware \emph{data valuation} in instruction tuning.

\paragraph{Our approach.}
We propose \framework, a gradient-based data selection framework utilizing a LoRA ensemble on a frozen small proxy backbone (i.e., GPT-2).
This method captures intrinsic dynamics by tracking per-example gradients across ensemble members and epochs and aggregates them into a \emph{Gradient Signal-to-Noise Ratio} (G-SNR), a metric that unifies \emph{learning progress} (via early-to-late gradient-drop statistics) and \emph{epistemic uncertainty} (via ensemble variance). Derived solely from the gradients of the training objective, G-SNR is \emph{objective-agnostic}—obviating the need for task-specific rewards or manual annotations. 
Finally, we rank examples by G-SNR to identify high-value subsets for fine-tuning target models.

Our contributions are fourfold:
\begin{enumerate}
\tightlist
\item \textbf{LoRA ensembles for uncertainty-aware data valuation.}
To the best of our knowledge, we are the first to leverage a LoRA ensemble to model epistemic uncertainty for gradient-based data valuation and selection in instruction tuning.
\item \textbf{\framework.}
We propose \framework, an efficient, objective-agnostic data selection method that runs a LoRA ensemble on a frozen LLM backbone with a small GPT-2 proxy and aggregates per-example gradients into a gradient signal-to-noise (G-SNR) utility.
\item \textbf{Strong empirical gains.}
On Alpaca and Alpaca-GPT4 with LLaMA-2-7B/13B, models fine-tuned on \framework-selected 5--15\% subsets match or outperform Random and Superfiltering in 19/24 LLM-as-a-judge evaluation cases, and a small human study confirms that these preferences align with human judgments.
\item \textbf{Faster convergence.}
Training on \framework-selected subsets converges faster than competitive filtering baselines under the same compute budget, without degrading final instruction-following quality.
\end{enumerate}

\section{Related Work}
\subsection{Data Valuation and Selection}

Estimating the importance of individual training examples is a long-standing problem in machine learning.
Influence-function methods approximate the effect of upweighting or removing a training point on the loss at a target point via second-order Taylor expansions~\citep{koh2017understanding} or first-order approximations such as TracIn~\citep{pruthi2020estimating}. 
While theoretically appealing, these approaches are prohibitively expensive and often brittle for deep networks, failing to scale cleanly to modern LLM fine-tuning pipelines.

Recent work adapts gradient-based data valuation to instruction tuning.
LESS~\citep{xia2024less} introduces an optimizer-aware influence formulation, explicitly modeling Adam updates to bulid a gradient datastore with low-rank (LoRA-based) representations for gradient similarity search.
This requires computing and storing per-example gradient features of a strong base model and tailoring selection to a particular downstream task, rather than exploiting the training dynamics of the fine-tuning run itself.

Superfiltering~\citep{li2024superfiltering} studies weak-to-strong data filtering for instruction tuning. 
It shows that small proxy models such as GPT-2 produce perplexity and instruction-following difficulty (IFD) rankings highly correlated with those of larger LLaMA-2 models, and uses the proxy to compute IFD scores and select a top-$k$ subset for fine-tuning the strong model. 
This substantially reduces filtering cost while often matching or surpassing full-data baselines, but assigns each example a static difficulty score from the pre-trained proxy, without modeling interactions between training samples or the evolution of gradients and epistemic uncertainty during fine-tuning.

\subsection{Ensembles and Uncertainty Modeling}

Epistemic uncertainty~\citep{amini2020deep} is particularly important in large-scale instruction tuning, where training data may be noisy or spurious and we would like to identify examples on which the model is most uncertain.
A classical line of work uses approximate Bayesian methods such as Monte Carlo dropout~\citep{gal2016dropout} or deep ensembles~\citep{lakshminarayanan2017simple} to quantify epistemic uncertainty via disagreement across multiple stochastic forward passes or independently trained models.
However, full ensembles of large language models are computationally expensive, limiting their practicality in standard fine-tuning pipelines.

In the era of LLMs, LoRA~\citep{hu2022lora} plays a central role by enabling parameter-efficient adaptation while keeping the backbone frozen.
LoRA-Ensemble~\citep{muhlematter2024lora} extends uncertainty quantification by training multiple independent LoRA adapters on a shared backbone, emulating a deep ensemble without duplicating the full model.
This strategy retains much of the accuracy and calibration benefits of deep ensembles at a fraction of the parameter and compute cost.

\paragraph{Positioning our approach.}
Motivated by these lines of work, we design \framework\ as an in-situ, gradient-based data selection method that combines a small proxy model (GPT-2) with a LoRA ensemble on a frozen LLM backbone.

\section{LoRA-Ensemble Approximation for Gradient-Based Data Valuation}
\label{sec:lora_ensemble}
\subsection{LoRA-Ensemble Approximation}

We employ a parameter-efficient ensemble based on LoRA adapters, where all ensemble members share a frozen backbone $\theta_0$, but each maintains its own low-rank adapter parameters $\Delta\theta^{(m)}$ for $m = 1,\dots,M$. For each self-attention layer, we attach rank-$r$ adapters to the query/key/value projections as in standard LoRA, and initialize the adapters of different ensemble members independently.

Let $\Delta\theta^{(m)}_e$ denote the LoRA parameters of member $m$ after $e$ epochs of fine-tuning, and
\begin{equation}
\theta^{(m)}_e \;=\; \theta_0 + \Delta\theta^{(m)}_e
\end{equation}
the full parameter vector. All members start from the same pretrained backbone with randomly initialized adapters, i.e., $\Delta\theta^{(m)}_0$ for $m=1,\dots,M$, and are fine-tuned on $\mathcal{D}$ for $T$ epochs with different random seeds or data orderings. This yields $M$ adapted models that share a common backbone but follow distinct parameter trajectories.
We denote by $g_i^{(m,e)}$ the per-example gradient for sample $i$ produced by LoRA member $m$ at epoch $e$ (formally defined in Equation~\ref{eq:grad}).
Under first-order optimization, small parameter updates—and hence local changes in loss around a given checkpoint—are approximately linear in these gradients. 
Our data valuation method therefore relies on the collection of per-example gradients $\{g_i^{(m,e)}\}_{m,e}$ as a compact summary of how $(x_i,y_i)$ drives the ensemble during training.

In practice, we train the ensemble by minimizing the average per-member loss on each batch, $\mathcal{L}_{\mathrm{ens}} = \tfrac{1}{M}\sum_{m=1}^M \mathcal{L}(f_{\theta^{(m)}_e}(x_i), y_i)$, and backpropagating this objective with respect to each adapter. This vectorized implementation is equivalent (up to a constant rescaling of the learning rate) to training $M$ independent LoRA models that reuse the same backbone computation. Crucially, we do \emph{not} optimize the loss of an averaged prediction, $\mathcal{L}(\tfrac{1}{M}\sum_m f_{\theta^{(m)}_e}(x_i), y_i)$, which would explicitly encourage all members to correct the same ensemble error and thus suppress the epistemic disagreement that our gradient-variance statistics are intended to capture.

\paragraph{Gradient-based view of LoRA ensembles.}
Our use of a LoRA ensemble relies on two structural properties of parameter-efficient fine-tuning. First, empirical studies suggest that LoRA and related adapters typically keep the fine-tuned model close to the pretrained backbone, so that a first-order Taylor expansion around $\theta_0$ provides an accurate local linearization of outputs and losses in the adapter subspace~\citep{li2025efficient}. In this locally linear regime, the per-example LoRA gradient $g_i^{(m,e)}$ is the direction of steepest loss change for member $m$ at epoch $e$, and its norm reflects how strongly $(x_i,y_i)$ drives parameter updates.

Second, independently initialized LoRA adapters $\{\Delta\theta^{(m)}\}_{m=1}^M$ can be interpreted as approximate samples from a posterior over low-rank updates around $\theta_0$~\citep{muhlematter2024lora,balabanov2024uncertainty}. Under this view, the collection $\{g_i^{(m,e)}\}_{m=1}^M$ approximates samples from a distribution over update directions induced by $(x_i,y_i)$ at epoch $e$: the ensemble mean estimates its expected update magnitude, and the ensemble variance quantifies epistemic uncertainty about how the model should adapt on this example. Our data valuation scheme builds on these two moments: we prioritize examples whose expected gradient is large in early epochs and decays over training, and we down-weight examples whose gradient variance remains high at late epochs, corresponding to regions where the ensemble fails to reach a stable consensus. These intuitions are instantiated in the uncertainty-aware gradient-drop score introduced in Section~\ref{sec:gsnr}.

\begin{figure}[t]
    \centering
    \begin{subfigure}[t]{0.49\columnwidth}
        \centering
        \includegraphics[width=\linewidth]{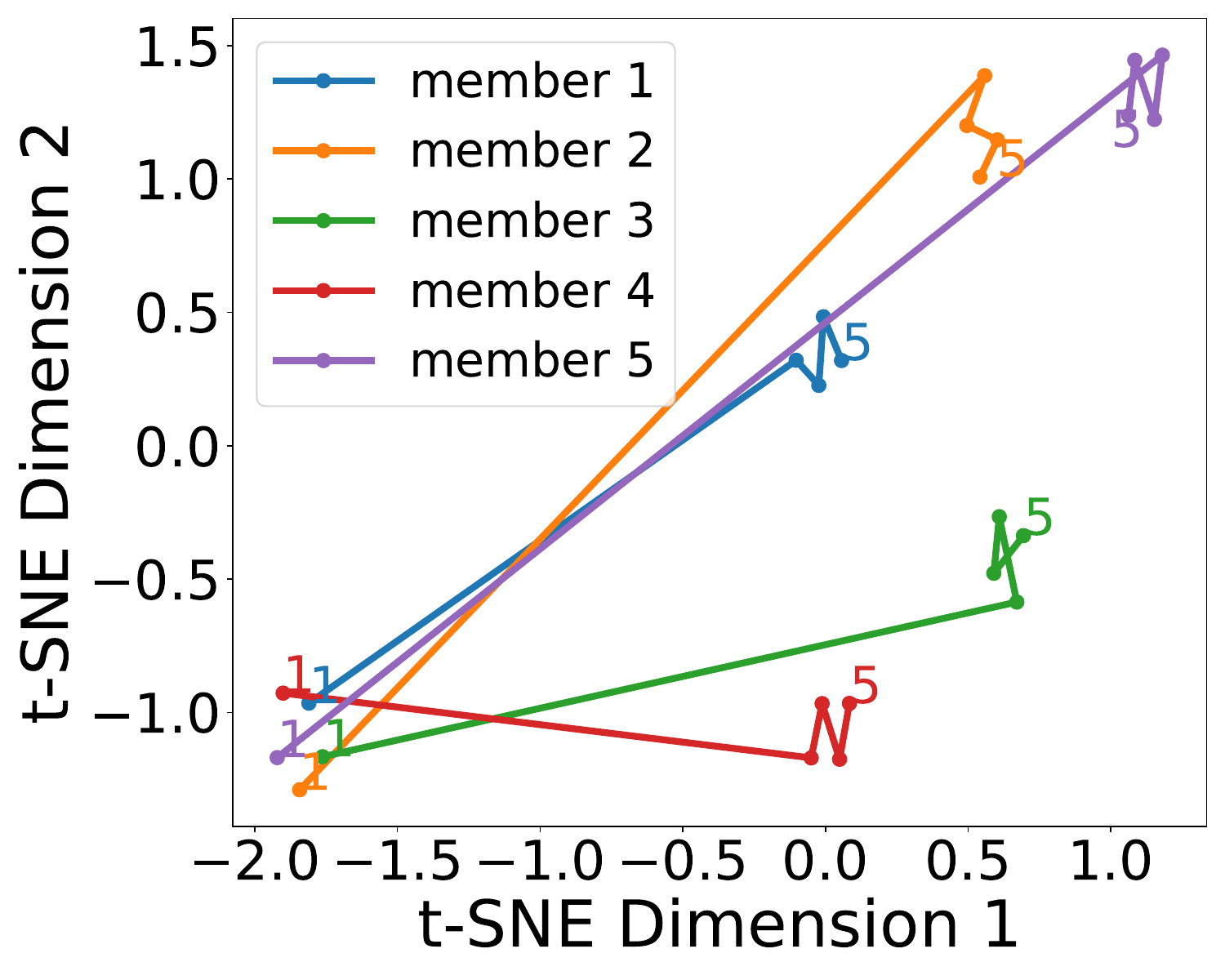}
        \caption{Alpaca trajectories with GPT-2 filter.}
        \label{fig:traj-alpaca}
    \end{subfigure}
    \hfill
    \begin{subfigure}[t]{0.49\columnwidth}
        \centering
        \includegraphics[width=\linewidth]{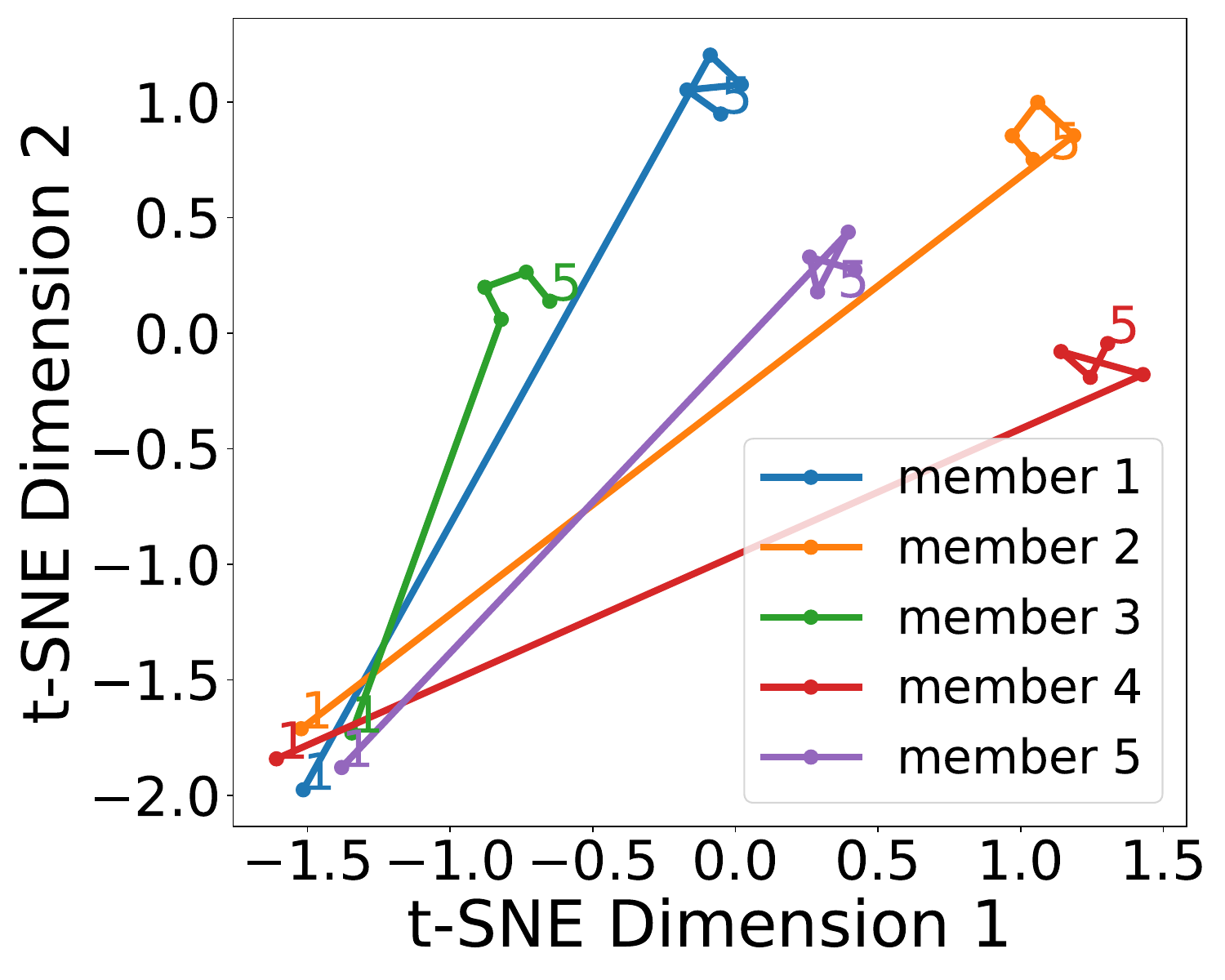}
        \caption{Alpaca-GPT4 trajectories with GPT-2 filter.}
        \label{fig:traj-alpaca-gpt4}
    \end{subfigure}
    \caption{
        Trajectories of LoRA ensemble members in a low-dimensional embedding of gradient profiles for (a) Alpaca and (b) Alpaca-GPT4. Each polyline corresponds to one LoRA member; markers along the line denote successive epochs.
    }
    \label{fig:traj-tsne}
\end{figure}

\paragraph{Empirical geometry of LoRA training dynamics.}
To empirically support this view, we visualize the trajectories of LoRA ensemble members in a low-dimensional embedding of their per-example gradient profiles (Figures~\ref{fig:traj-alpaca} and~\ref{fig:traj-alpaca-gpt4}). For each LoRA member $m$ and epoch $e$, we construct a high-dimensional feature vector $\mathbf{g}_{m,e} \in \mathbb{R}^N$ whose $i$-th coordinate encodes the relative gradient drop on example $i$ between epoch $1$ and epoch $e$. We stack all $\{\mathbf{g}_{m,e}\}_{m,e}$, apply PCA to reduce the dimensionality from $N$ to a small $d$ (e.g., $d=25$) for denoising and stabilization, and then run t-SNE on the PCA embeddings to obtain two-dimensional points $\mathbf{z}_{m,e} \in \mathbb{R}^2$ in a shared “gradient-profile space.” For each member $m$, we connect $\mathbf{z}_{m,1} \rightarrow \mathbf{z}_{m,2} \rightarrow \cdots \rightarrow \mathbf{z}_{m,T}$ to form a trajectory; this should be interpreted as mapping each epoch to a global pattern of relative gradient drops across all training examples for that member, rather than as a trajectory of individual examples.

Empirically, each polyline starts from a similar region of the embedding but quickly diverges, with different members moving toward distinct areas and then following stable, non-collapsing trajectories. At the same time, later-epoch points are more tightly clustered than early-epoch points, indicating that ensemble disagreement is gradually reduced as training proceeds. These patterns provide qualitative evidence that (i) LoRA adapters induce a rich, structured distribution over update directions around the frozen backbone, and (ii) the mean and variance of per-example gradients across members capture non-degenerate epistemic uncertainty, thereby justifying our use of these statistics for data valuation.

\subsection{Practical Considerations}
\label{sec:prac_consideration}
In practice, we set the ensemble size to $M=5$. As shown in Figure~\ref{fig:traj-tsne}, the trajectories of the LoRA ensemble members change substantially between epochs 1 and 2, and then remain relatively stable afterwards.
Motivated by this observation, we set the terminal epoch to be $T=2$ when computing gradient statistics, which further reduces computational cost while preserving the essential training-dynamics signal.

\section{\framework\ : Data Selection with G-SNR}
\begin{figure*}[t]
  \centering
  \includegraphics[width=\textwidth]{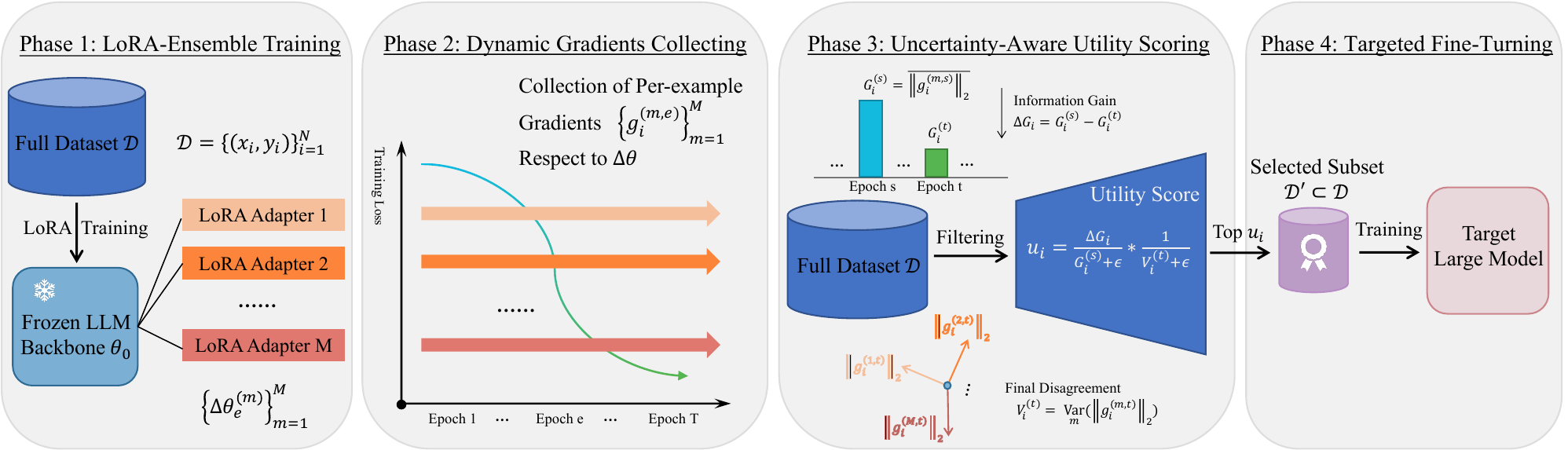}
  \caption{
  Overview of our gradient-based data selection pipeline.
  \textbf{Phase 1} trains a set of LoRA-ensemble members that share the same frozen backbone $\theta_0$ on the full dataset $\mathcal{D}$, producing multiple adapters $\{\Delta\theta_e^{(m)}\}_{m=1}^M$ after $e$ epochs.
  \textbf{Phase 2} collects \emph{per-example} gradients respect to $\Delta\theta$ during training process at different epochs, yielding gradient profiles $\{g_i^{(m,e)}\}_{m=1}^M$.
  \textbf{Phase 3} calculates an uncertainty-aware utility score based on these profiles by combining (i) an \emph{information-gain} term $\Delta G_i = G_i^{(s)} - G_i^{(t)}$ that measures gradient drop from early to later training, and (ii) a \emph{disagreement} term $V_i^{(t)}=\mathop{\mathrm{Var}}\limits_{m}\!\big(\|g_i^{(m,t)}\|_2\big)$ that captures ensemble variability, via an SNR-style form $u_i \propto \frac{\Delta G_i}{G_i^{(s)}+\epsilon}\cdot\frac{1}{V_i^{(t)}+\epsilon}$.
  \textbf{Phase 4} selects the top-ranked subset $\mathcal{D}'\subset\mathcal{D}$ based on the utility score $u_i$ for training the targeted model.
  }
  \label{fig:framework}
\end{figure*}

Figure~\ref{fig:framework} summarizes our gradient-based data selection pipeline \framework. Section~\ref{sec:lora_ensemble} establishes that a LoRA ensemble provides a rich, non-degenerate latent space, justifying Phases 1–2 in Figure~\ref{fig:framework}.
In this section, we focus on Phase~3 and derive our Gradient Signal-to-Noise Ratio (G-SNR) utility, which turns ensemble training dynamics into data-selection scores.

\subsection{Problem Setup}

Let $\mathcal{D} = \{(x_i, y_i)\}_{i=1}^N$ be a large-scale instruction--response dataset, where $x_i$ is an instruction and $y_i$ its reference response. We are given a pretrained language model and wish to fine-tune it for instruction following. Our goal is to assign each training example $(x_i, y_i)$ a scalar utility score $u_i$ and to use these scores to select a subset $\mathcal{D}' \subset \mathcal{D}$ (or to reweight $\mathcal{D}$) such that fine-tuning on $\mathcal{D}'$ achieves comparable or better performance than using the full dataset.

\subsection{Per-Example Gradient Statistics}

We characterize each training example by how difficult it is initially, how much the ensemble learns from it, and how stable its effect is across ensemble members, using its per-example LoRA gradients. Let $T$ be the total number of training epochs. For example $i$ in ensemble member $m$ at epoch $e \in \{1,\dots,T\}$ we define
\begin{equation}
\label{eq:grad}
g_i^{(m,e)} = \nabla_{\Delta\theta^{(m)}_e}
\,\mathcal{L}\big(f_{\theta^{(m)}_e}(x_i), y_i\big),
\end{equation}
where $\mathcal{L}$ is the token-level training loss and gradients are taken only with respect to the LoRA adapters $\Delta\theta^{(m)}_e$. Thus $g_i^{(m,e)}$ directly measures how strongly $(x_i,y_i)$ pushes the adapted parameters to change at a given epoch.

\paragraph{Gradient magnitude.}
For each example $i$ and epoch $e$, we summarize its instantaneous training signal by the ensemble-averaged LoRA-gradient norm
\begin{equation}
   G_i^{(e)} \;=\; \frac{1}{M} \sum_{m=1}^{M} \big\| g_i^{(m,e)} \big\|_2, 
\end{equation}
where $g_i^{(m,e)}$ is the per-example gradient of member $m$ at epoch $e$.
Larger $G_i^{(e)}$ indicates that $(x_i,y_i)$ induces a stronger parameter update at that stage of training, reflecting its current difficulty or influence.

\paragraph{Ensemble disagreement via gradient variance.}
To quantify how consistently the ensemble reacts to each example, we measure the variance of gradient norms across ensemble members at each epoch,
\begin{equation}
V_i^{(e)} =
\frac{1}{M} \sum_{m=1}^{M} \big\| g_i^{(m,e)} \big\|_2^2
-
\left(
\frac{1}{M} \sum_{m=1}^{M} \big\| g_i^{(m,e)} \big\|_2
\right)^2
\end{equation}
$V_i^{(e)}$ serves as a stability indicator: low values indicate that ensemble members agree on the update magnitude for $(x_i,y_i)$, while high values signals persistent disagreement and potential ambiguity or noise.

\subsection{Gradient Signal-to-Noise Ratio (G-SNR) for Data Selection}
\label{sec:gsnr}

We propose a \textbf{Gradient Signal-to-Noise Ratio (G-SNR)} utility to select high-quality training examples from a large candidate pool. The key motivation is a global optimization view: because all samples share parameters, the utility of a data point should reflect not only its local training signal but also how \emph{consistently} this signal aligns with the overall descent direction across different optimization trajectories.

\paragraph{Gradient-drop signal.}
We measure an example's learning progress using a \emph{gradient-drop} signal computed between an early training stage and a later stage. Concretely, we train an ensemble of $M$ lightweight adapters (e.g., LoRA) and record per-example gradient-norm statistics at two time points: 
an early epoch $s$ and a later epoch $t$ ($s < t$). Let 
$G_i^{(m,s)}$ and $G_i^{(m,t)}$ denote the gradient norm of sample $i$ under ensemble member $m$ at epochs $s$ and $t$, and we define the ensemble means $G_i^{(s)} = \tfrac{1}{M}\sum_{m=1}^{M} G_i^{(m,s)}$ and $G_i^{(t)} = \tfrac{1}{M}\sum_{m=1}^{M} G_i^{(m,t)}$, as well as the later-stage variance $V_i^{(t)} = \tfrac{1}{M}\sum_{m=1}^{M}\big(G_i^{(m,t)}\big)^2 - \big(G_i^{(t)}\big)^2$. The raw drop $\Delta_i = G_i^{(s)} - G_i^{(t)}$ captures how much the sample's gradient magnitude decreases as training proceeds:
intuitively, examples that induce stable and effective descent tend to exhibit a clearer reduction.

\paragraph{Uncertainty normalization (G-SNR).}
To account for uncertainty in the gradient signal, we down-weight examples whose gradient statistics are highly variable across ensemble members. 
Our G-SNR utility is
\begin{equation}
u_i^{\textsc{G-SNR}}
=
\frac{G_i^{(s)} - G_i^{(t)}}{G_i^{(s)} + \epsilon}
\cdot
\frac{1}{V_i^{(t)} + \epsilon},
\label{eq:gsnr}
\end{equation}
where $\epsilon$ is a small constant for numerical stability. 
The first factor is a \emph{relative} gradient drop that normalizes out scale effects (so that large-gradient examples are not trivially favored), while the second factor penalizes high-variance, unreliable signals by down-weighting examples whose late-stage gradients disagree across ensemble members. 
In this sense, G-SNR behaves like a signal-to-noise ratio: it prefers large, consistent gradient drops (signal) while suppressing examples with high ensemble disagreement (noise).

\paragraph{Selection protocol.}
Given utilities $\{u_i^{\textsc{G-SNR}}\}_{i=1}^N$, we select the top-$\alpha$ fraction of examples as the training subset. 
We use the same computation and selection procedure across datasets and base models, making G-SNR objective-agnostic: it only assumes access to per-example gradients under a parametric loss, without relying on a specific instruction template or external proxy score.

\section{Experiments}
\label{sec:experiments}

\paragraph{Overview.}
We empirically evaluate \framework\ as a gradient-based data selection method for instruction tuning along three axes.
First, on small subsets (5--15\%), we ask whether models fine-tuned on \framework-selected data match or outperform random splits and the strong baseline Superfiltering in pairwise preference against full-data models under identical training setups.
Second, we ablate alternative gradient-based utilities to examine their behavior and confirm that our uncertainty-aware, normalized G-SNR formulation is the most robust across datasets, base models, and adaptation regimes.
Third, from an optimization standpoint, we compare training-loss trajectories and convergence, showing that \framework-selected subsets yield faster convergence than competitive filtering baselines under the same compute budget.

\subsection{Experiments Setup}
\label{subsec:exp_setup}

\paragraph{Datasets.}
We train all models on two standard instruction-tuning corpora, Alpaca~\cite{taori2023alpaca} and Alpaca-GPT4~\cite{peng2023instruction}, each containing 52{,}000 instruction–response pairs.
For evaluation, we follow common practice and use the WizardLM evaluation set~\cite{xu2023wizardlm} (218 instructions) and the Vicuna evaluation set~\cite{zheng2023judging} (80 instructions), both consisting of open-ended instruction-following prompts designed for pairwise preference judgments.

\paragraph{Implementation details.}
We instantiate \framework on two backbone models from the LLaMA-2 family~\cite{touvron2023llama}: LLaMA-2-7B and LLaMA-2-13B.
For each backbone, we consider both LoRA-based adaptation (low-rank adapters on a frozen backbone) and full-parameter fine-tuning, and we keep the training protocol (data, number of epochs, and optimization settings) fixed across different data selection strategies to ensure a fair comparison.
All models are optimized with Adam~\cite{kingma2014adam}, using a learning rate of $2\times10^{-5}$ for LLaMA-2-7B and $1\times10^{-5}$ for LLaMA-2-13B.
For the GPT-2 proxy used to compute G-SNR, we apply LoRA with rank $r{=}8$ and $\alpha{=}16$, train with a learning rate of $5\times10^{-5}$, and record per-example gradients at epochs 1 and 2 as the initial and terminal references, respectively, as justified in Section~\ref{sec:prac_consideration}.

\paragraph{Evaluation setting.} We adopt the LLM-as-a-judge paradigm to compare, in a pairwise manner, the model fine-tuned on selected data against the full-data model, using the prompt template from Vicuna~\cite{zheng2023judging}, detailed in Appendix~\ref{appendix:eval}.
For evaluation, we use GPT-5.1 and Qwen3-235B-Instruct as our proprietary and open-source judge models, respectively, and report scores averaged over both; the open-source judge is included to facilitate replication in case proprietary APIs become unavailable in the future.
We further validate that LLM-judge preferences with five human annotators.

\subsection{Main Results}
\label{subsec:main_results}

\begin{table*}[t]
    \centering
    \small
    \setlength{\tabcolsep}{3pt}
    \begin{tabular}{l*{16}{c}}
        \toprule
        \multirow{2}{*}{Dataset / Base Model} &
        \grouphead{Alpaca\\LLaMA2-7B} &
        \grouphead{Alpaca\\LLaMA2-13B} &
        \grouphead{Alpaca-GPT4\\LLaMA2-7B} &
        \grouphead{Alpaca-GPT4\\LLaMA2-13B} \\
        \cmidrule(lr){2-5}
        \cmidrule(lr){6-9}
        \cmidrule(lr){10-13}
        \cmidrule(lr){14-17}
        & 5\% & 10\% & 15\% & 100\%
        & 5\% & 10\% & 15\% & 100\%
        & 5\% & 10\% & 15\% & 100\%
        & 5\% & 10\% & 15\% & 100\% \\
        \midrule
        Random Split (LoRA)
        & 0.97 & 0.93 & 0.92 & 1.00
        & 0.96 & 0.99 & 0.96 & 1.00
        & 0.91 & 0.89 & 0.98 & 1.00
        & 0.90 & 0.91 & 0.91 & 1.00 \\

        Superfiltering (LoRA)
        & \textbf{1.02} & \textbf{1.10} & 1.02 & 1.00
        & 1.18 & 1.10 & 1.11 & 1.00
        & 0.94 & \textbf{1.12} & 1.13 & 1.00
        & 0.98 & 1.01 & 1.00 & 1.00 \\

        Ours (LoRA)
        & 1.01 & 1.07 & \textbf{1.06} & 1.00
        & \textbf{1.19} & \textbf{1.20} & \textbf{1.12} & 1.00
        & \textbf{1.04} & 1.02 & \textbf{1.15} & 1.00
        & \textbf{1.05} & \textbf{1.03} & \textbf{1.10} & 1.00 \\
        \midrule
        Random Split (Full)
        & 0.96 & 0.95 & 0.98 & 1.00
        & 0.94 & 0.98 & 0.94 & 1.00
        & 0.92 & 0.97 & 0.99 & 1.00
        & 0.94 & 0.93 & 0.97 & 1.00 \\

        Superfiltering (Full)
        & 1.05 & 1.08 & 1.11 & 1.00
        & 1.01 & 1.09 & 1.05 & 1.00
        & \textbf{1.15} & 1.04 & 1.03 & 1.00
        & 1.00 & 1.12 & \textbf{1.15} & 1.00 \\

        Ours (Full)
        & \textbf{1.16} & \textbf{1.16} & \textbf{1.12} & 1.00
        & \textbf{1.09} & \textbf{1.10} & \textbf{1.11} & 1.00
        & 1.01 & \textbf{1.06} & \textbf{1.05} & 1.00
        & \textbf{1.08} & \textbf{1.13} & 1.01 & 1.00 \\
        \bottomrule
    \end{tabular}
    \caption{
        Pairwise winning scores for instruction tuning.
        Columns group results for Alpaca and Alpaca-GPT4 fine-tuning on LLaMA2-7B and LLaMA2-13B, each using 5\%, 10\%, 15\%, or 100\% of the corresponding training set.
        Rows compare three data selection strategies (Random Split, Superfiltering, and Ours) under two adaptation regimes: LoRA (low-rank adapters on a frozen backbone) and Full (full-parameter fine-tuning).
        The Pairwise Winning Score (PWS) is computed as $1 + \bigl(\mathrm{Num(Win)} - \mathrm{Num(Lose)}\bigr) / \mathrm{Num(All)}$.
        The consistent improvements across the evaluation benchmarks demonstrate the effectiveness of \framework.
    }
    \label{tab:inst_filtering_main}
\end{table*}

Table~\ref{tab:inst_filtering_main} reports Pairwise Winning Scores (PWS)~\citep{li2024superfiltering} for Alpaca and Alpaca-GPT4 across backbones (7B/13B), subset ratios (5\%/10\%/15\%), and adaptation regimes (LoRA/Full).
Overall, \framework\ consistently outperforms random subsets and is better or comparable with current SOTA data filtering baseline Superfiltering \cite{li2024superfiltering}.

\paragraph{Comparison to Random.}
Random subset selection is frequently below the full-data baseline (PWS$<1$), especially at smaller ratios.
In contrast, \framework\ is much more robust: in many settings, models trained on only 5\%--15\% of the data match or exceed the full-data counterpart, indicating that the selected examples preserve high training signal.

\paragraph{Comparison to Superfiltering.}
Superfiltering is a strong baseline and can be favorable in some configurations.
However, \framework\ provides more consistent gains across backbones and regimes, with particularly strong improvements in several 13B settings and in full-parameter fine-tuning where optimization dynamics and sample interactions are more pronounced.
These results support the central hypothesis of \framework: incorporating global training dynamics and gradient uncertainty yields a stronger signal for selecting effective instruction-tuning examples than single-score heuristics alone.

\subsection{Human Evaluation}
\label{subsec:human_eval}

We compare LLaMA2-13B (full fine-tuning) trained on 10\% data against the 100\% baseline for Alpaca and Alpaca-GPT4, using 100 prompts sampled from WizardLM\cite{xu2023wizardlm} and Vicuna\cite{zheng2023judging}, with the criteria same as the previous pair-wise
evaluation, i.e. Helpfulness, Relevance, Accuracy,
and Level of Detail. The win/tie/lose counts are 44/19/37 (Alpaca) and 49/8/43 (Alpaca-GPT4), consistent with the LLM-judge trend, supporting that Table~\ref{tab:inst_filtering_main} reflects perceived quality.

\subsection{Ablation Study}
\label{subsec:ablation}

\newcommand{\groupheadthree}[1]{%
  \multicolumn{3}{c}{\begin{tabular}{@{}c@{}}#1\end{tabular}}%
}

\begin{table*}[t]
    \centering
    \small
    \setlength{\tabcolsep}{3pt}
    \begin{tabular}{l*{12}{c}}
        \toprule
        \multirow{2}{*}{Utility / Base Model} &
        \groupheadthree{Alpaca\\LLaMA2-7B} &
        \groupheadthree{Alpaca\\LLaMA2-13B} &
        \groupheadthree{Alpaca-GPT4\\LLaMA2-7B} &
        \groupheadthree{Alpaca-GPT4\\LLaMA2-13B} \\
        \cmidrule(lr){2-4}
        \cmidrule(lr){5-7}
        \cmidrule(lr){8-10}
        \cmidrule(lr){11-13}
        & 5\% & 10\% & 15\%
        & 5\% & 10\% & 15\%
        & 5\% & 10\% & 15\%
        & 5\% & 10\% & 15\% \\
        \midrule
        Utility Alternative \#1 (LoRA)
        & -0.04 & -0.19 & -0.21
        & -0.04 & +0.04 & -0.03
        & -0.44 & -0.33 & -0.32
        & -0.52 & -0.23 & -0.23 \\

        Utility Alternative \#2 (LoRA)
        & -0.07 & -0.14 & -0.17
        & -0.06 & -0.20 & -0.14
        & -0.37 & -0.29 & -0.33
        & -0.29 & -0.11 & -0.21 \\

        Utility Alternative \#3 (LoRA)
        & -0.09 & +0.01 & -0.16
        & -0.05 & -0.11 & -0.10
        & -0.35 & -0.19 & -0.34
        & -0.23 & -0.14 & -0.25 \\
        \midrule
        Utility Alternative \#1 (Full)
        & -0.31 & -0.31 & -0.22
        & -0.04 & +0.00 & -0.18
        & -0.41 & -0.37 & -0.20
        & -0.22 & -0.14 & -0.02 \\

        Utility Alternative \#2 (Full)
        & -0.23 & -0.23 & -0.13
        & -0.07 & -0.13 & -0.22
        & -0.18 & -0.34 & -0.28
        & -0.15 & -0.25 & -0.18 \\

        Utility Alternative \#3 (Full)
        & -0.16 & -0.11 & -0.02
        & -0.02 & -0.10 & -0.13
        & -0.10 & -0.13 & -0.41
        & -0.08 & -0.20 & -0.23 \\
        \bottomrule
    \end{tabular}
    \caption{
    Ablation on gradient-based utility functions for data selection.
    Columns follow the same configurations as Table~\ref{tab:inst_filtering_main} (Alpaca / Alpaca-GPT4 on LLaMA2-7B and LLaMA2-13B with 5\%, 10\%, and 15\% of the training set).
    Rows compare three utility alternatives under LoRA and Full, where entries report the gain/drop in pairwise winning score relative to our default utility in Table~\ref{tab:inst_filtering_main}.
    Utility Alternative~\#1 uses $G_1 - G_T$, Alternative~\#2 uses $(G_1 - G_T)/(G_1 + \varepsilon)$, Alternative~\#3 uses $(G_1 - G_T)/(V_T + \varepsilon)$, and our G-SNR is $((G_1 - G_T)/(G_1 + \varepsilon)) \cdot (1/(V_T + \varepsilon))$ (See Equation~\ref{eq:gsnr}).
    }
    \label{tab:inst_utility_ablation}
\end{table*}

Table~\ref{tab:inst_utility_ablation} ablates the design of the gradient-based utility by replacing our default G-SNR (Eq.~\ref{eq:gsnr}) with three simpler variants, and reports the resulting gain or drop in pairwise winning score relative to the \framework\ results in Table~\ref{tab:inst_filtering_main}.

\paragraph{Why normalization and uncertainty matter.}
Across datasets, model sizes, and adaptation regimes, all three alternatives typically yield negative deltas, indicating that none of them matches the robustness of G-SNR.
Using only the raw gradient drop ($G_1 - G_T$, Alternative~\#1) performs worst overall, suggesting that absolute scale alone is not a reliable indicator of data utility.
Normalizing by the initial gradient magnitude (Alternative~\#2) or by the late-stage variance (Alternative~\#3) mitigates some of this instability but still consistently underperforms the full G-SNR formulation.
These trends support our design choice: combining a relative gradient-drop term with an uncertainty-based penalty provides a more stable and informative signal than either component in isolation.

\paragraph{Negative utilities.}
Because $G_1 - G_T$ can be negative, G-SNR and its variants can also produce negative utilities. 
In practice, this simply pushes such examples toward the bottom of the ranking; under our top-$\alpha$ selection rule, only the relative ordering matters.
Empirically, downweighting highly negative utilities is desirable, as they often correlate with examples whose gradients do not exhibit consistent improvement over training.

\subsection{Convergence Analysis}
\label{subsec:convergence}


\begin{figure}[t]
  \centering
  \begin{subfigure}[t]{0.49\columnwidth}
    \centering
    \includegraphics[width=\linewidth]{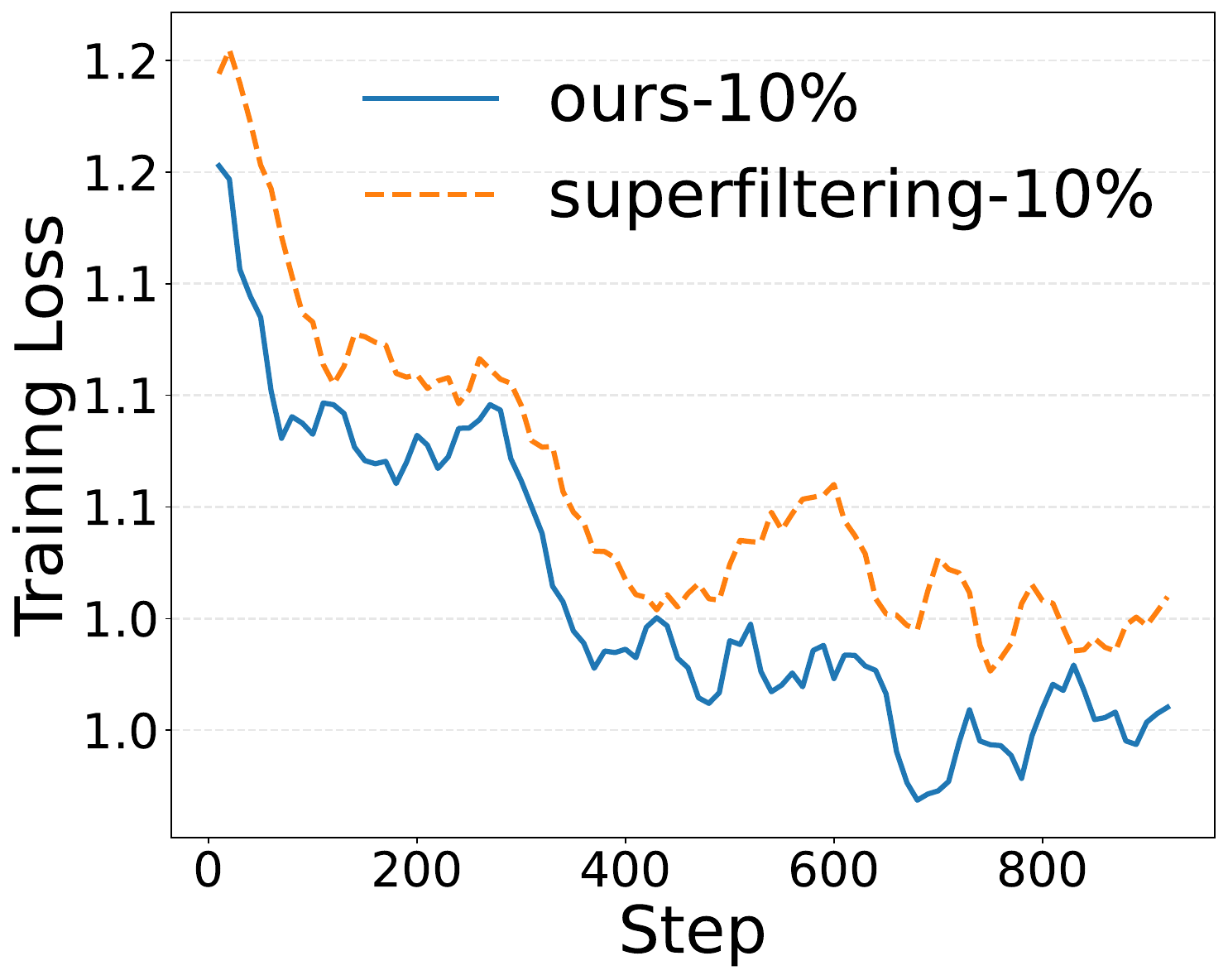}
    \caption{Full-parameter.}
    \label{fig:trainloss-13b-alpaca-full-10}
  \end{subfigure}
  \hfill
  \begin{subfigure}[t]{0.49\columnwidth}
    \centering
    \includegraphics[width=\linewidth]{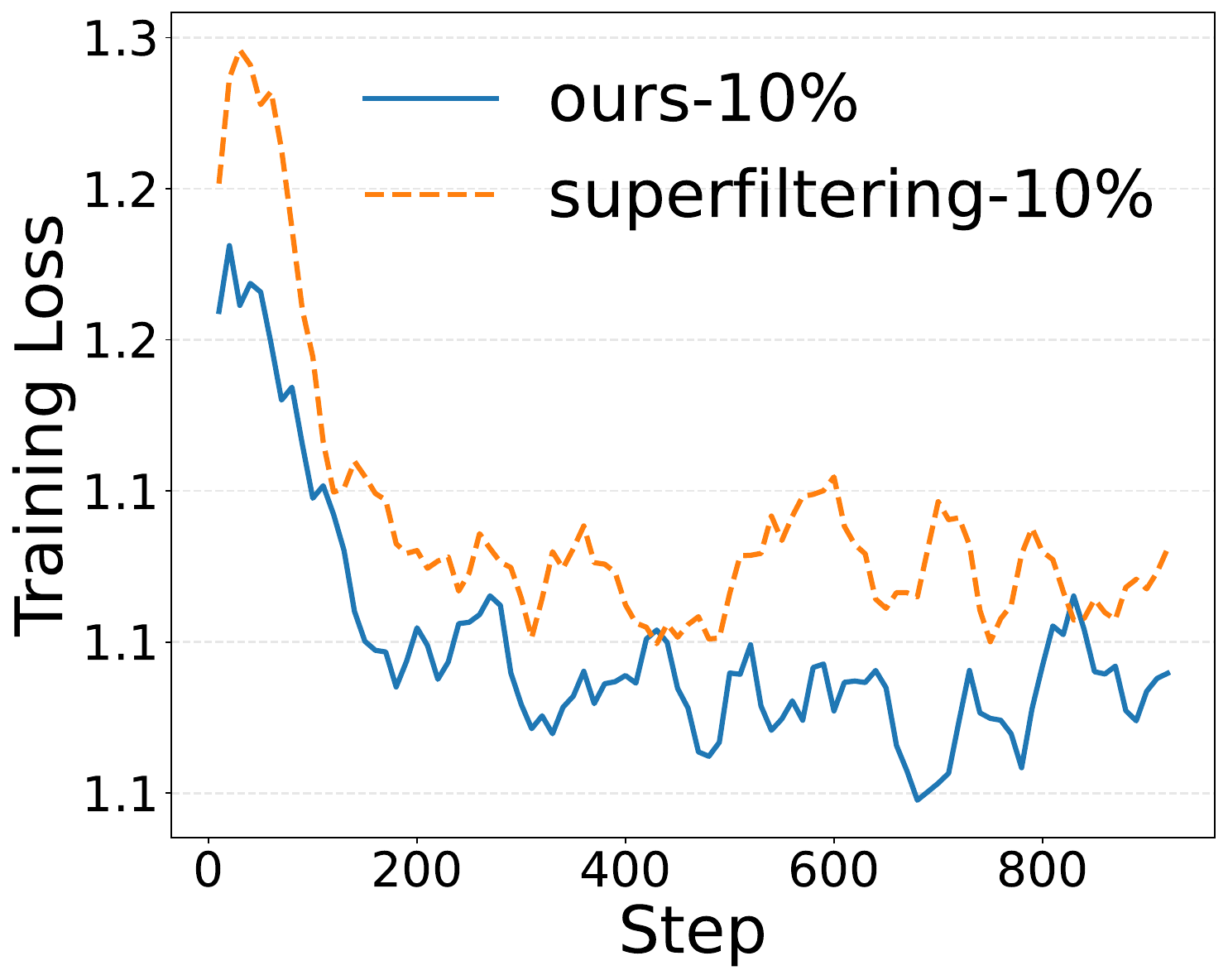}
    \caption{LoRA.}
    \label{fig:trainloss-13b-alpaca-lora-10}
  \end{subfigure}

  \caption{
    Training loss for LLaMA-2-13B on Alpaca with 10\% of the data.
    Left: full-parameter fine-tuning; right: LoRA-based fine-tuning.
    In both settings, we compare convergence speed and final training loss
    between our selection (ours-10\%) and Superfiltering-10\%.
  }
  \label{fig:trainloss-13b-alpaca-10}
\end{figure}

Beyond preference scores, \framework\ is motivated by a global optimization view: examples interact through shared parameters, so a good subset should drive stable, efficient descent. Figure~\ref{fig:trainloss-13b-alpaca-10} compares training-loss curves for Superfiltering and \framework\ under identical settings (LLaMA-2-13B, 10\% Alpaca as examples), for both full fine-tuning and LoRA. In both regimes, \framework\ converges faster and reaches lower loss earlier, suggesting our uncertainty-aware, gradient-based criterion better matches the global training dynamics than optimizing a single proxy. 
While \framework\ could be suspected to favor “easy” data, Table~\ref{tab:inst_filtering_main} shows consistent preference gains over strong baselines, indicating improved response quality rather than a trivial bias.

\section{Further Discussion}
\paragraph{What does G-SNR capture?}
Our experiments suggest that G-SNR is a useful proxy for data utility in instruction tuning: examples with large, consistent gradient drops under the LoRA ensemble tend to yield stronger downstream instruction-following performance than those with small or noisy gradient changes. Conceptually, G-SNR can be viewed as a gradient-based analogue of dataset cartography: instead of tracking per-example loss trajectories, it summarizes how strongly an example pulls the model in parameter space between early and late training, and how stable this effect is across ensemble members. 

\paragraph{Objective-agnostic and model-agnostic.}
Because G-SNR is computed purely from per-example gradients of a parametric loss, it is \emph{objective-agnostic}: it does not require task-specific reward models, preference labels, or handcrafted difficulty scores, and can in principle be applied to any training objective for which gradients are available. Likewise, \framework\ is largely \emph{model-agnostic}: beyond requiring a differentiable backbone and a parameter-efficient adaptation mechanism (here, LoRA on LLaMA-2 with a GPT-2 proxy), the framework does not assume a particular architecture or instruction format. Due to computational constraints and page limits, we instantiate and evaluate \framework\ only on supervised instruction tuning, but we argue that the same G-SNR principle can naturally extend to other objectives (e.g., multi-task mixtures), which we leave for future work.

\section{Conclusion}
We presented \textsc{GradFiltering}, a gradient-based data selection framework for instruction tuning. The method fine-tunes a small GPT-2 proxy with a LoRA ensemble, records per-example adapter gradients during training, and aggregates them into a Gradient Signal-to-Noise Ratio (G-SNR) utility that combines relative gradient drop with late-epoch gradient variance. Because it is defined purely in terms of per-example gradients, G-SNR is objective-agnostic and can, in principle, be paired with a wide range of training objectives.

Empirically, \textsc{GradFiltering} delivers strong and consistent gains. On Alpaca and Alpaca-GPT4 with LLaMA-2-7B/13B under both LoRA and full-parameter fine-tuning, models trained on \textsc{GradFiltering}-selected 5--15\% subsets match or outperform random splits and the strong Superfiltering baseline in most LLM-as-a-judge settings, with a small human study confirming the same preference trends. Moreover, \textsc{GradFiltering}-selected subsets converge faster and reach lower training loss than competitive filters under the same compute budget, supporting our view that uncertainty-aware gradient statistics provide an effective signal for curating large instruction-tuning corpora.

\section*{Limitations}

Despite its empirical robustness, \framework\ has several limitations. First, it operates on gradient norms and their variance, ignoring gradient \emph{direction}; examples that are locally uninformative in norm but crucial for aligning with rare or long-horizon behaviors may be under-valued. Second, G-SNR relies on a specific early/late snapshot scheme and a modest ensemble size ($M=5$); although our t-SNE analyses indicate that these settings already induce diverse trajectories, different training schedules or proxies could change the behavior. Third, the proxy model must still be fine-tuned with backpropagation, which is cheaper than running a strong teacher or constructing a full gradient datastore, but not free. Finally, like other training-dynamics methods, G-SNR assumes that “useful” examples look useful early in training; regimes with delayed credit assignment or strong curriculum effects may violate this assumption.

\section*{Acknowledgments}

\bibliography{gradfiltering_source/main}

\clearpage
\appendix

\section{Appendix}
\label{sec:appendix}

\subsection{Prompt for Evaluation}
\label{appendix:eval}

We follow the LLM-as-a-judge protocol described in the main text, using a fixed, symmetric prompt to compare responses from two candidate models on the same instruction.
For each evaluation example, the judge model (GPT-5.1 or Qwen3-235B-Instruct) is adopted.
We adopt the prompt template from Vicuna~\cite{zheng2023judging}, with only minimal renaming of model identifiers.

\begin{table}[t]
\centering

\begin{tcolorbox}[
  enhanced,
  colback=white,
  colframe=black,
  boxrule=0.8pt,
  left=8pt,right=8pt,top=8pt,bottom=8pt,
  arc=2mm,
  width=\linewidth,
]
\centering
{\large\bfseries Prompt for Performance Evaluation}\par
\vspace{6pt}
\hrule
\vspace{10pt}

\RaggedRight
\textbf{System Prompt}\par
\vspace{2pt}
You are a helpful and precise assistant for checking the quality of the answer.\par

\vspace{10pt}
\textbf{User Prompt}\par
\vspace{2pt}
[Question]\par
\textit{Question}\par
[The Start of Assistant 1's Answer]\par
\textit{Answer 1}\par
[The End of Assistant 1's Answer]\par
[The Start of Assistant 2's Answer]\par
\textit{Answer 2}\par
[The End of Assistant 2's Answer]\par

\vspace{10pt}
We would like to request your feedback on the performance of two AI assistants in response to the user question displayed above.
Please rate the helpfulness, relevance, accuracy, level of details of their responses.
Each assistant receives an overall score on a scale of 1 to 10, where a higher score indicates better overall performance.
Please first output a single line containing only two values indicating the scores for Assistant 1 and 2, respectively.
The two scores are separated by a space.
In the subsequent line, please provide a comprehensive explanation of your evaluation, avoiding any potential bias and ensuring that the order in which the responses were presented does not affect your judgment.\par
\end{tcolorbox}

\caption{Prompt template used for LLM-as-a-judge evaluation.
The judge model sees the instruction and two anonymized candidate responses (\texttt{Model A} and \texttt{Model B}) and is asked to output a preference label (\texttt{A}, \texttt{B}, or \texttt{Tie}) together with a brief explanation.
To avoid ordering bias, we shuffle the order of \texttt{A} and \texttt{B}.}
\label{tab:judge_prompt_template}

\end{table}
\end{document}